# Dempster-Shafer Theory for Move Prediction in Start Kicking of The Bicycle Kick of Sepak Takraw Game

Andino Maseleno, Md. Mahmud Hasan

*Computer Science Program, Universiti Brunei Darussalam, Brunei BE1410*

## Abstract

This paper presents Dempster-Shafer theory for move prediction in start kicking of the bicycle kick of sepak takraw game. Sepak takraw is a highly complex net-barrier kicking sport that involves dazzling displays of quick reflexes, acrobatic twists, turns and swerves of the agile human body movement. A Bicycle kick or Scissor kick is a physical move made by throwing the body up into the air, making a shearing movement with the legs to get one leg in front of the other without holding on to the ground. Specifically, this paper considers bicycle kick of sepak takraw game in start kicking of the ball with uncertainty where player has different awareness regarding the contingencies. We have chosen Dempster-Shafer theory because the advantages of the Dempster-Shafer theory which include the ability to model information in a flexible way without requiring a probability to be assigned to each element in a set, providing a convenient and simple mechanism for combining two or more pieces of evidence under certain conditions, it can model ignorance explicitly, rejection of the law of additivity for belief in disjoint propositions.

## INTRODUCTION

Sepak takraw is a spectacular three-a-side game in which a ball is propelled over a high net using any part of the body other than the hands-usually the foot, knee, shoulder, or head. The game combines soccer and gymnastics. Play begins with the server standing in the service circle with his or her teammates in the quarter circles. On the other side, one player has to have a foot in the service circle, but the others can stand anywhere. A player in the quarter circle tosses the ball to the server, who sends it over the net. As in volleyball, each side can strike the ball three times before it returns to the opposition half [1]. In this paper we consider Dempster-Shafer theory for move prediction in start kicking of the bicycle kick of sepak takraw game.

Some research related with human movement have been developed which were prediction of the movement patterns for human squat jumping using the inverse-inverse dynamics technique [2] and neural network for human arm movement prediction in CVEs [3]. The movement patterns for human squat jumping using the inverse-inverse dynamics technique, in this study, Inverse-Inverse dynamics was used to predict the movement patterns in human vertical jumping. Inverse-inverse dynamic can, just as optimal control strategies, be subjected to optimization to determine the optimum motion. The novelty of the method lies in the choice of independent variables. While optimum control strategies traditionally use muscle

---

**\*Corresponding Author: Email: andinomaseleno@pasca.gadjahmada.edu,**

Phone + 6738896460

activation or joint moment variations, inverse-inverse dynamics parameterizes the movement and treats joint moments or muscle activation as dependent variables. This means that if the motion is partially unknown, some parametric functions can be assumed and the optimization algorithm based on the results of the inverse simulation identifies the unknown parameters. In other words, the joint angles of the musculoskeletal model are found using optimization and the forces or moments needed for motion are calculated using inverse dynamics [2].

Neural network for human arm movement prediction in CVEs, the purpose of this experiment was to assess the quality of feed forward back propagation neural networks in predicting natural avatar arm movement used in a CVE. In addition the experiment attempted to find the bounds for precise neural network prediction. The results show many different combinations of back propagation neural network topologies are capable of predicting up to 400 ms of human arm movements relatively accurately [3].

Stern et al. [4] investigate the problem of learning to predict moves in the board game of Go from game records of expert players. This method has two major components: a) a pattern extraction scheme for efficiently harvesting patterns of given size and shape from expert game records and b) a Bayesian learning algorithm (in two variants) that learns a distribution over the values of a move given a board position based on the local pattern context. The system is trained on 181,000 expert games and shows excellent prediction performance as indicated by its ability to perfectly predict the moves made by professional Go players in 34% of test positions.

Monte Carlo search, and specifically the Upper Confidence Bounds applied to Trees algorithm, has contributed to a significant improvement in the game of Go and has received considerable attention in other applications [5], this article investigates two enhancements to the Upper Confidence Bounds applied to Trees algorithm. First, it consider the possible adjustments to Upper Confidence Bounds applied to Trees when the search tree is treated as a graph (and information amongst transpositions are shared). The second modification introduces move groupings, which may reduce the effective branching factor. Experiments with both enhancements were performed using artificial trees and in the game of Go. From the experimental results we conclude that both exploiting the graph structure and grouping moves may contribute to an increase in the playing strength of game programs using Upper Confidence Bounds applied to Trees.

The advantages of the Dempster-Shafer theory which include the ability to model information in a flexible way without requiring a probability to be assigned to each element in a set, providing a convenient and simple mechanism for combining two or more pieces of evidence under certain conditions, it can model ignorance explicitly, rejection of the law of additivity for belief in disjoint propositions. We propose that progress can be made if we adopt a unified, rigorous and consistent mechanism for representing and managing the uncertainty that results from the complexity of the bicycle kick. Specifically, this paper considers bicycle kick of sepak takraw game in start kicking of the ball with uncertainty where player has different awareness regarding the contingencies.

**SEPAK TAKRAW GAME**

Sepak takraw is a skill ball game, which requires the use of the feet and head to keep the ball in the air and in a targeted direction. Sepak takraw or kick volleyball is a sport native to Southeast Asia, resembling volleyball, except that it uses a rattan ball and only allows players to use their feet and head to touch the ball. A cross between football and volleyball, it is a popular sport in Thailand, Cambodia, Malaysia, Laos, Philippines and Indonesia. Lately, the game becomes famous outside Asia such as United States, Canada, England, Germany, Brazil and New Zealand where the game played as recreational activity and tournament. The garneplay described briefly as spiking a ball into the opponent court to achieve point. The strategies in sepak takraw are also very similar to those in volleyball. The receiving team will attempt to play the takraw ball towards the front of the net, making the best use of their 3 hits, to set and spike the ball [6]. Figure 1 shows bicycle kick of sepak takraw game and figure 2 shows start kicking of bicycle kick . We describe ten motions based on hypothesis of sepak takraw player to start kicking of bicycle kick.

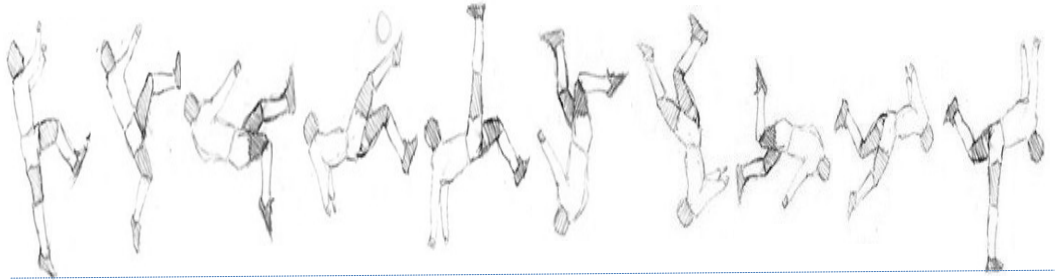

Fig. 1: Bicycle kick of sepak takraw game

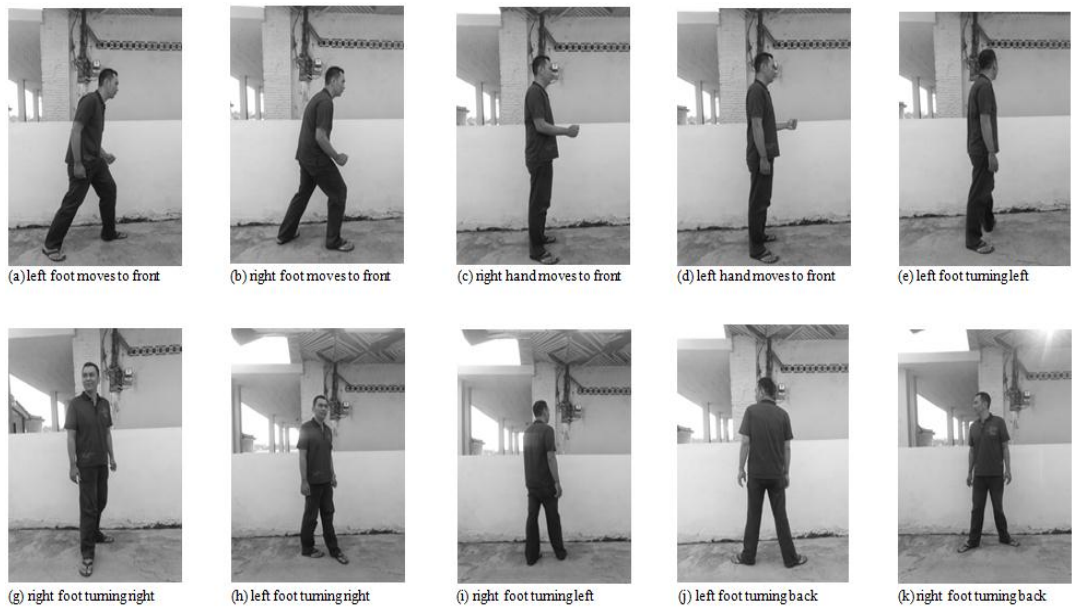

Fig. 2: Start kicking of bicycle kick

**DEMPSTER-SHAFER THEORY**

The Dempster-Shafer theory [7] or the theory of belief functions is a mathematical theory of evidence which can be interpreted as a generalization of probability theory in which the elements of the sample space to which nonzero probability mass is attributed are not single points but sets [8]. The sets that get nonzero mass are called focal elements. The sum of these probability masses is one, however, the basic difference between Dempster-Shafer theory and traditional probability theory is that the focal elements of a Dempster-Shafer structure may overlap one another. The Dempster-Shafer theory also provides methods to represent and combine weights of evidence.

Considering a finite set (frame of discernment) $\Theta$, a basic probability assignment is a function $m: 2^\Theta \rightarrow [0, 1]$ so that $m(\emptyset) = 0$, $A \subseteq \Theta$ $m(A) = 1$ and $m(A) \geq 0$ for all $A \subseteq \Theta$. The subsets of $\Theta$ which are associated with nonzero values of $m$ are known as focal elements and the union of the focal elements is called core. The value of $m(A)$ expresses the proportion of all relevant and available evidence that supports the claim that a particular element of $\Theta$ belongs to the set $A$ but to no particular subset of $A$. This value pertains only to the set $A$ and makes no additional claims about any subsets of $A$. From this kind of mass assignment, the upper and lower bounds of a probability interval can be defined. Shafer defined the concepts of belief and plausibility as two measures over the subsets of $\Theta$ as follows.

$$\text{Bel}(A) = \sum_{B \subseteq A} m(B) \tag{1}$$

$$\text{Pl}(A) = \sum_{B \cap A \neq \emptyset} m(B) \tag{2}$$

A basic probability assignment can also be viewed as determining a set of probability distributions $P$ over $\Theta$ so that $\text{Bel}(A) \leq P(A) \leq \text{Pl}(A)$. It can be easily seen that these two measures are related to each other as $\text{Pl}(A) = 1 - \text{Bel}()$. Therefore, one needs to know only one of the three values of $m$, Bel, or Pl to derive the other two. Dempster's rule of combination can be used for pooling of evidence from two belief functions Bel1 and Bel2 over the same frame of discernment, but induced by different independent sources of information. The Dempster's rule of combination for combining two sets of masses, $m1$ and $m2$ is defined as follows.

$$m_{12}(\emptyset) = 0 \tag{3}$$

$$m_{12}(A) = \frac{1}{1-k} \sum_{B \cap C = A \neq \emptyset} m_1(B) m_2(C) \tag{4}$$

$$k = \sum_{B \cap C = \emptyset} m_1(B) m_2(C) \tag{5}$$

Here $k$ is a measure of the amount of conflict between two evidences. If $k = 1$ the two evidences cannot be combined because their cores are disjoint. This rule is commutative, associative, but not idempotent or continuous.

**IMPLEMENTATION**

As shown in figure 2, we describe ten motions which include left foot moves to front, right foot moves to front, right hand moves to front, left hand moves to front, left foot turning left, right foot turning left, left foot turning right, right foot turning right, left foot turning back and right foot turning back. The strategy followed in the Dempster-Shafer theory for dealing with uncertainty roughly amounts to starting with an initial set of hypotheses. Basic probability assignments for each motion as shown in table 1, the values for basic probability assignment are determined based on hypothesis, suppose we have nine different conditions.

Table 1: Basic probability assignment

| No. | Motion | Moving | Basic Probability Assignment | | | | | | | | |
|---|---|---|---|---|---|---|---|---|---|---|---|
| | | | Condition 1 | Condition 2 | Condition 3 | Condition 4 | Condition 5 | Condition 6 | Condition 7 | Condition 8 | Condition 9 |
| 1 | Left foot moves to front | Front | 0.75 | 0.55 | 0.55 | 0.55 | 0.45 | 0.45 | 0.45 | 0.45 | 0.65 |
| 2 | Right foot moves to front | Front | 0.75 | 0.75 | 0.55 | 0.45 | 0.45 | 0.45 | 0.45 | 0.65 | 0.65 |
| 3 | Right hand moves to front | Front | 0.55 | 0.55 | 0.45 | 0.45 | 0.45 | 0.45 | 0.65 | 0.65 | 0.75 |
| 4 | Left hand moves to front | Front | 0.55 | 0.45 | 0.45 | 0.45 | 0.45 | 0.65 | 0.65 | 0.75 | 0.75 |
| 5 | Left foot turning left | Left Back | 0.45 | 0.45 | 0.45 | 0.45 | 0.65 | 0.65 | 0.75 | 0.75 | 0.55 |
| 6 | Right foot turning left | Left Back | 0.45 | 0.45 | 0.45 | 0.65 | 0.65 | 0.75 | 0.75 | 0.55 | 0.55 |
| 7 | Left foot turning right | Right Back | 0.45 | 0.45 | 0.65 | 0.65 | 0.75 | 0.75 | 0.55 | 0.55 | 0.45 |
| 8 | Right foot turning right | right Back | 0.45 | 0.65 | 0.65 | 0.75 | 0.75 | 0.55 | 0.55 | 0.45 | 0.45 |
| 9 | Left foot turning back | back | 0.65 | 0.65 | 0.75 | 0.75 | 0.55 | 0.55 | 0.45 | 0.45 | 0.45 |
| 10 | Right foot turning back | back | 0.65 | 0.75 | 0.75 | 0.55 | 0.55 | 0.45 | 0.45 | 0.45 | 0.45 |

*4.1 Motion 1*

Left foot moves to front {F}

$m_1\{F\} = 0.75$

$m_1\{\Theta\} = 1 - 0.75 = 0.25$

*4.2 Motion 2*

We combine two motions which include left foot moves to front and right foot moves to front as shown in table 2.

Table 2: The first combination

| | | {F} | 0.75 | Θ | 0.25 |
|---|---|---|---|---|---|
| {F} | 0.75 | {F} | 0.56 | {F} | 0.19 |
| Θ | 0.25 | {F} | 0.19 | Θ | 0.06 |

*4.3 Motion 3*

We combine three motions which include left foot moves to front, right foot moves to front and right hand moves to front as shown in table 3.

Table 3: The second combination

| | | {F} | 0.55 | Θ | 0.45 |
|---|---|---|---|---|---|
| {F} | 0.94 | {F} | 0.52 | {F} | 0.42 |
| Θ | 0.06 | {F} | 0.03 | Θ | 0.03 |

*4.4 Motion 4*

We combine four motions which include left foot moves to front, right foot moves to front, right hand moves to front and left hand moves to front as shown in table 4.

Table 4: The third combination

| | | {F} | 0.55 | Θ | 0.45 |
|---|---|---|---|---|---|
| {F} | 0.97 | {F} | 0.53 | {F} | 0.44 |
| Θ | 0.03 | {F} | 0.02 | Θ | 0.01 |

*4.5 Motion 5*

We combine five motions which include left foot moves to front, right foot moves to front, right hand moves to front, left hand moves to front and left foot turning left as shown in table 5.

Table 5: The fourth combination

|     |      |       |      |     |      |
|-----|------|-------|------|-----|------|
|     |      | {L, B}| 0.45 | Θ   | 0.55 |
| {F} | 0.99 | { Θ } | 0.45 | {F} | 0.54 |
| Θ   | 0.02 | {L, B}| 0.01 | Θ   | 0.01 |

*4.6 Motion 6*

We combine six motions which include left foot moves to front, right foot moves to front, right hand moves to front, left hand moves to front, left foot turning left and right foot turning left as shown in table 6.

Table 6: The fifth combination

|       |      |        |      |        |      |
|-------|------|--------|------|--------|------|
|       |      | {L, B} | 0.45 | Θ      | 0.55 |
| {F}   | 0.98 | { Θ }  | 0.44 | {F}    | 0.54 |
| {L, B}| 0.02 | {L, B} | 0.01 | {L, B} | 0.01 |
| Θ     | 0.02 | {L, B} | 0.01 | Θ      | 0.01 |

*4.7 Motion 7*

We combine seven motions which include left foot moves to front, right foot moves to front, right hand moves to front, left hand moves to front, left foot turning left, right foot turning left and left foot turning right as shown in table 7.

Table 7: The sixth combination

|       |      |        |      |        |      |
|-------|------|--------|------|--------|------|
|       |      | {R, B} | 0.45 | Θ      | 0.55 |
| {F}   | 0.96 | { Θ }  | 0.43 | {F}    | 0.53 |
| {L, B}| 0.05 | {B}    | 0.02 | {L, B} | 0.03 |
| Θ     | 0.02 | {R, B} | 0.01 | Θ      | 0.01 |

*4.8 Motion 8*

We combine eight motions which include left foot moves to front, right foot moves to front, right hand moves to front, left hand moves to front, left foot turning left, right foot turning left, left foot turning right and right foot turning right as shown in table 8.

Table 8: The seventh combination

|        |      |        |      |        |      |
|--------|------|--------|------|--------|------|
|        |      | {R, B} | 0.45 | Θ      | 0.55 |
| {F}    | 0.93 | Θ      | 0.42 | {F}    | 0.51 |
| {B}    | 0.04 | {B}    | 0.02 | {B}    | 0.02 |
| {L, B} | 0.05 | { B }  | 0.02 | {L, B} | 0.03 |
| {R, B} | 0.02 | {R, B} | 0.01 | {R, B} | 0.01 |
| Θ      | 0.02 | {R, B} | 0.01 | Θ      | 0.01 |

*4.9 Motion 9*

We combine nine motions which include left foot moves to front, right foot moves to front, right hand moves to front, left hand moves to front, left foot turning left, right foot turning left, left foot turning right, right foot turning right and left foot turning back as shown in table 9.

Table 9: The eighth combination

|        |      | {B} | 0.65 | Θ      | 0.35 |
|--------|------|-----|------|--------|------|
| {F}    | 0.88 | Θ   | 0.57 | {F}    | 0.31 |
| {B}    | 0.10 | {B} | 0.07 | {B}    | 0.04 |
| {L, B} | 0.05 | {B} | 0.03 | {L, B} | 0.02 |
| {R, B} | 0.05 | {B} | 0.03 | {R, B} | 0.02 |
| Θ      | 0.02 | {B} | 0.01 | Θ      | 0.01 |

*4.10 Motion 10*

We combine ten motions which include left foot moves to front, right foot moves to front, right hand moves to front, left hand moves to front, left foot turning left, right foot turning left, left foot turning right, right foot turning right, left foot turning back and right foot turning back as shown in table 10.

Table 10: The ninth combination

|        |      | {B} | 0.65 | Θ      | 0.35 |
|--------|------|-----|------|--------|------|
| {F}    | 0.72 | Θ   | 0.47 | {F}    | 0.25 |
| {B}    | 0.42 | {B} | 0.27 | {B}    | 0.15 |
| {L, B} | 0.05 | {B} | 0.03 | {L, B} | 0.02 |
| {R, B} | 0.05 | {B} | 0.03 | {R, B} | 0.02 |
| Θ      | 0.02 | {B} | 0.01 | Θ      | 0.01 |

The highest bpa value is the $m_{17}$ (B) that is equal to 0.92, it means the possibility of movement from ten motions which include left foot moves to front, right foot moves to front, right hand moves to front, left hand moves to front, left foot turning left, right foot turning left, left foot turning right, right foot turning right, left foot turning back and right foot turning back is moves to back.

**RESULT**

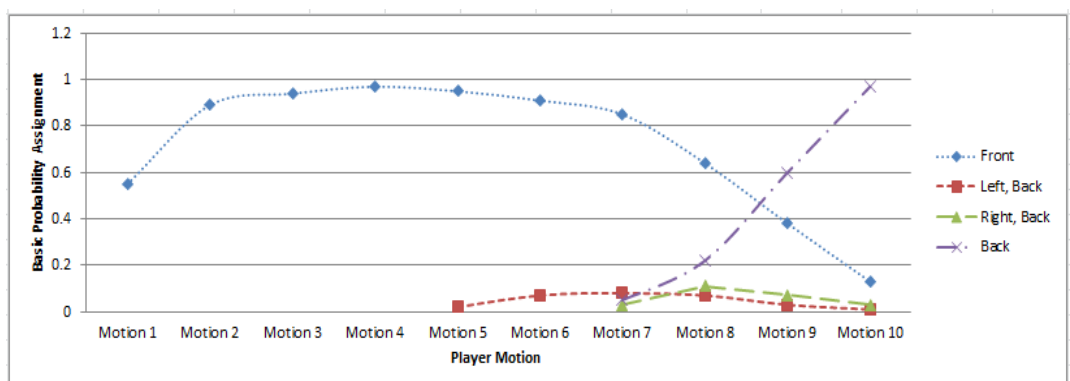

Fig. 3: condition 1

Figure 3 shows the graphic of condition 1, we get the highest basic probability assignment is back that is equal to 0.9, it shows from the last calculation of Dempster-Shafer on motion 10. It means the possibility of a player movements from ten motions which include left foot moves to front, right foot moves to front, right hand moves to front, left hand moves to front, left foot turning left, right foot turning left, left foot turning right, right foot turning right, left foot turning back, and right foot turning back is moves to back.

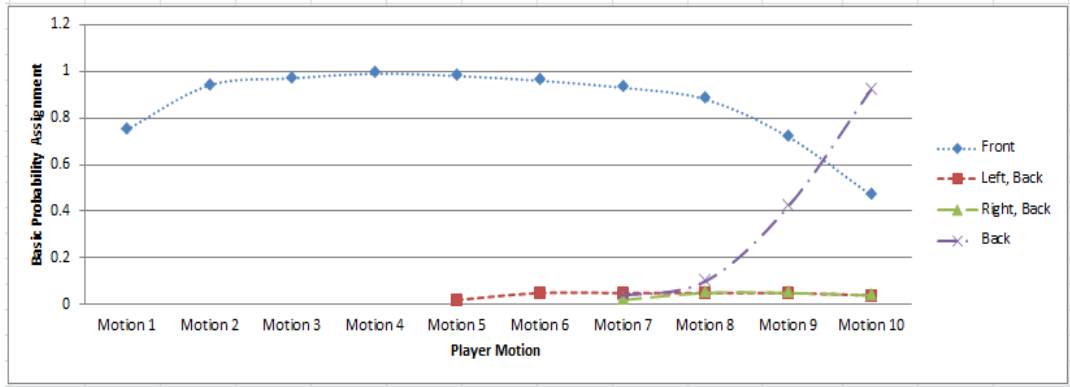

Fig. 4: condition 2

Figure 4 shows the graphic of condition 2, we get the highest basic probability assignment is back that is equal to 0.97, it shows from the last calculation of Dempster-Shafer on motion 10. It means the possibility of a player movements from ten motions which include left foot moves to front, right foot moves to front, right hand moves to front, left hand moves to front, left foot turning left, right foot turning left, left foot turning right, right foot turning right, left foot turning back, right foot turning back is moves to back.

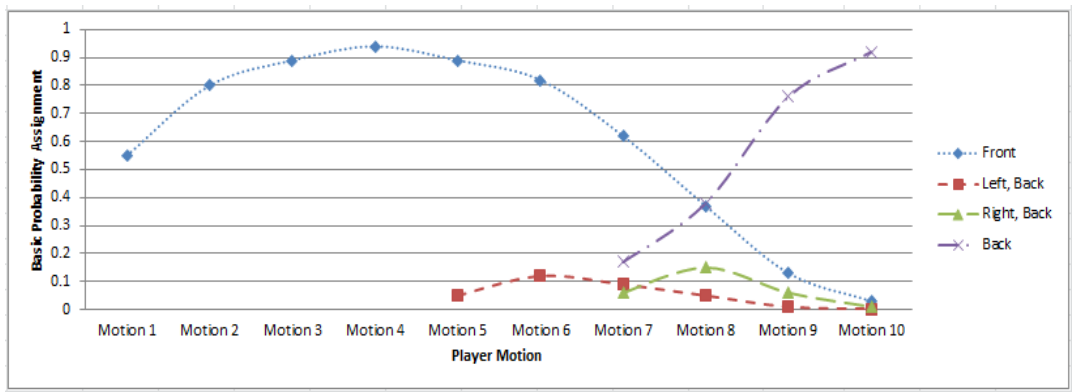

Fig. 5: condition 3

Figure 5 shows the graphic of condition 3, we get the highest basic probability assignment is back that is equal to 0.92, it shows from the last calculation of Dempster-Shafer on motion 10. It means the possibility of a player movements from ten motions which include left foot moves to front, right foot moves to front, right hand moves to front, left hand moves to front, left foot turning left, right foot turning left, left foot turning right, right foot turning right, left foot turning back, right foot turning back is moves to back.

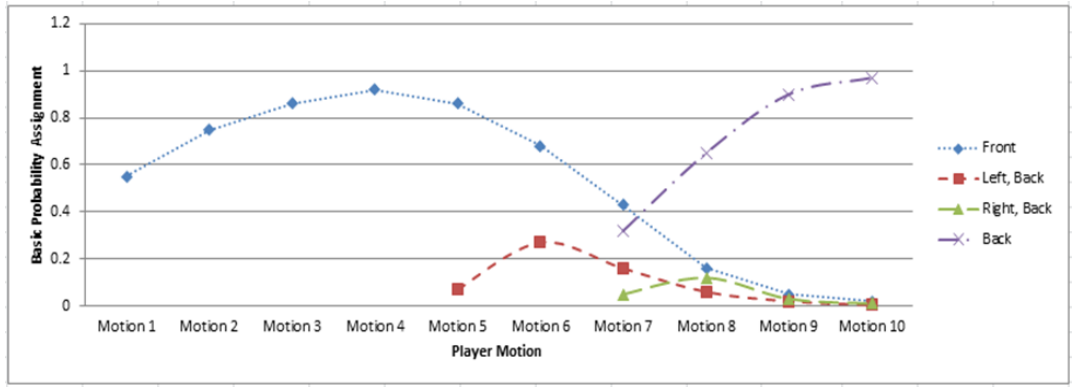

Fig. 6: Condition 4

Figure 6 shows the graphic of condition 4, we get the highest basic probability assignment is back that is equal to 0.97, it shows from the last calculation of Dempster-Shafer on motion 10. It means the possibility of a player movements from ten motions which include left foot moves to front, right foot moves to front, right hand moves to front, left hand moves to front, left foot turning left, right foot turning left, left foot turning right, right foot turning right, left foot turning back, right foot turning back is moves to back.

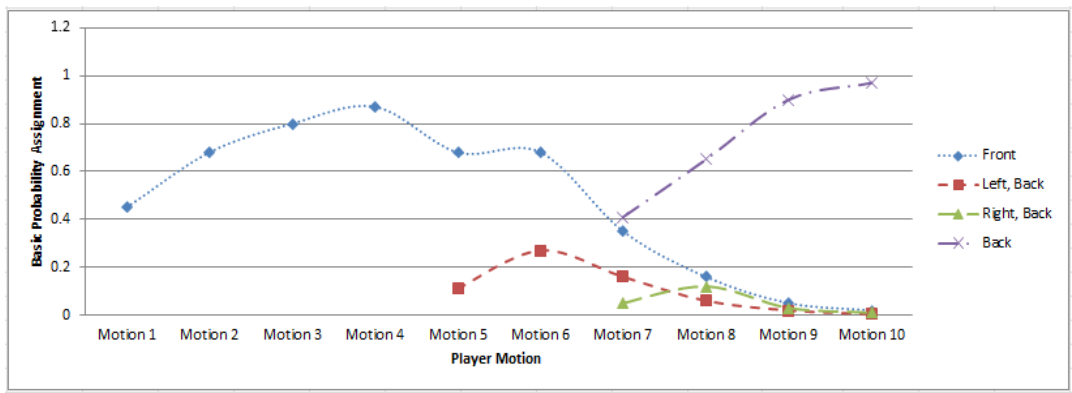

Fig. 7: Condition 5

Figure 7 shows the graphic of condition 5, we get the highest basic probability assignment is back that is equal to 0.97, it shows from the last calculation of Dempster-Shafer on motion 10. It means the possibility of a player movements from ten motions which include left foot moves to front, right foot moves to front, right hand moves to front, left hand moves to front, left foot turning left, right foot turning left, left foot turning right, right foot turning right, left foot turning back, right foot turning back is moves to back.

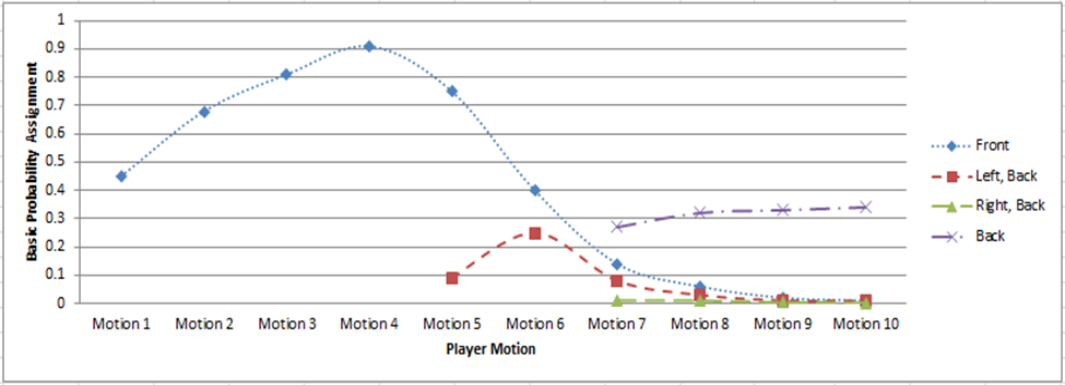

Fig. 8: Condition 6

Figure 8 shows the graphic of condition 6, we get the highest basic probability assignment is back that is equal to 0.34, it shows from the last calculation of Dempster-Shafer on motion 10. It means the possibility of a player movements from ten motions which include left foot moves to front, right foot moves to front, right hand moves to front, left hand moves to front, left foot turning left, right foot turning left, left foot turning right, right foot turning right, left foot turning back, right foot turning back is moves to back.

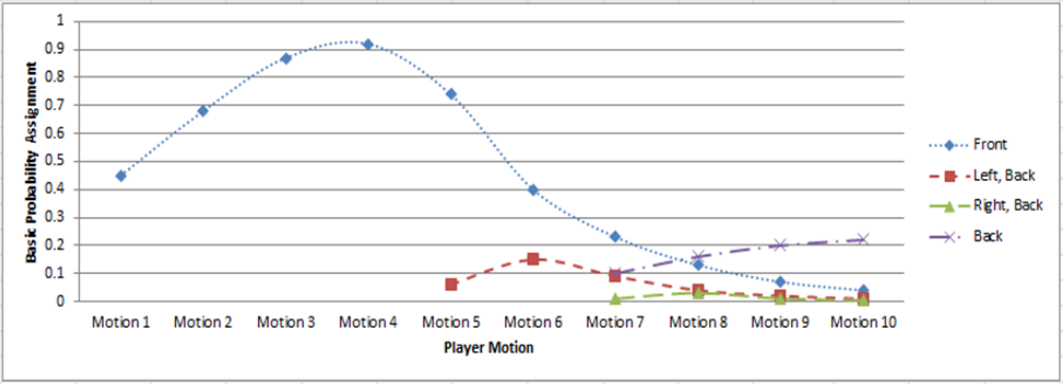

Fig. 9: Condition 7

Figure 9 shows the graphic of condition 7, we get the highest basic probability assignment is back that is equal to 0.22, it shows from the last calculation of Dempster-Shafer on motion 10. It means the possibility of a player movements from ten motions which include left foot moves to front, right foot moves to front, right hand moves to front, left hand moves to front, left foot turning left, right foot turning left, left foot turning right, right foot turning right, left foot turning back, right foot turning back is moves to back.

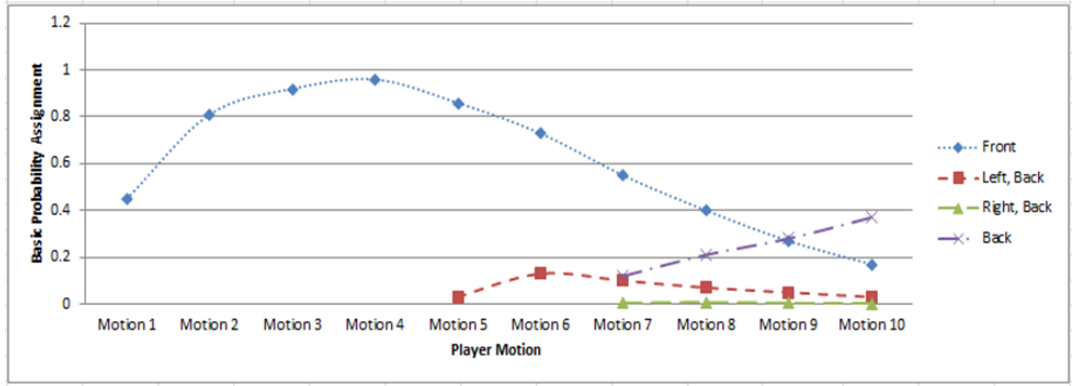

Fig. 10: Condition 8

Figure 10 shows the graphic of condition 8, we get the highest basic probability assignment is back that is equal to 0.37, it shows from the last calculation of Dempster-Shafer on motion 10. It means the possibility of a player movements from ten motions which include left foot moves to front, right foot moves to front, right hand moves to front, left hand moves to front, left foot turning left, right foot turning left, left foot turning right, right foot turning right, left foot turning back, right foot turning back is moves to back.

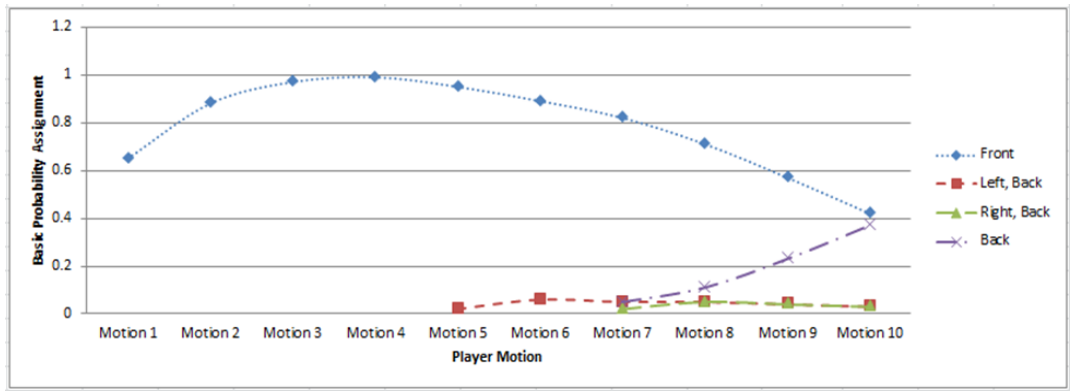

Fig. 11: Condition 9

Figure 11 shows the graphic of condition 9, we get the highest basic probability assignment is front that is equal to 0.49, it shows from the last calculation of Dempster-Shafer on motion 10. It means the possibility of a player movements from ten motions which include left foot moves to front, right foot moves to front, right hand moves to front, left hand moves to front, left foot turning left, right foot turning left, left foot turning right, right foot turning right, left foot turning back, right foot turning back is moves to front.

**CONCLUSION**

We proposed Dempster-Shafer theory for move prediction in start kicking of the bicycle kick of sepak takraw game. In this paper we describe ten motions which include left foot moves to front, right foot moves to front, right hand moves to front, left hand moves to front, left foot turning left, right foot turning left, left foot turning right, right foot turning right, left foot turning back and right foot turning back. The simplest possible method for using probabilities to quantify the uncertainty in a database is that of attaching a probability to every member of a relation, and to use these values to provide the probability that a particular value is the correct answer to a particular query. The knowledge is uncertain in the collection of basic events can be directly used to draw conclusions in simple cases, however, in many cases the various events associated with each other. Reasoning under uncertainty that used some of mathematical expressions, gave them a different interpretation: each piece of evidence may support a subset containing several hypotheses. This is a generalization of the pure probabilistic framework in which every finding corresponds to a value of a variable. In particular this experiment has shown Dempster-Shafer theory as an effective solution for player movement prediction in start kicking of the bicycle kick of sepak takraw game. As future works, we consider an extension of Dempster-Shafer theory with another uncertainty method.